# An Implementation of Vector Quantization using the Genetic Algorithm Approach

Maha Mohammed Khan
Department of Electrical Engineering
and Computer Science
University of Cincinnati
Cincinnati, Ohio-45219, USA
mahamofu@mail.uc.edu

*Abstract*— The application of machine learning(ML) and genetic programming(GP) to the image compression domain has produced promising results in many cases. The need for compression arises due to the exorbitant size of data shared on the internet. Compression is required for text, videos, or images, which are used almost everywhere on web be it news articles, social media posts, blogs, educational platforms, medical domain, government services, and many other websites, need packets for transmission and hence compression is necessary to avoid overwhelming the network. This paper discusses some of the implementations of image compression algorithms that use techniques such as Artificial Neural Networks, Residual Learning, Fuzzy Neural Networks, Convolutional Neural Nets, Deep Learning, Genetic Algorithms. The paper also describes an implementation of Vector Quantization using GA to generate codebook which is used for Lossy image compression. All these approaches prove to be very contrasting to the standard approaches to processing images due to the highly parallel and computationally extensive nature of machine learning algorithms. Such non-linear abilities of ML and GP make it widely popular for use in multiple domains. Traditional approaches are also combined with artificially intelligent systems, leading to hybrid systems, to achieve better results.

## I. Introduction

Data compression is a process that is performed on almost all forms of data, whether it be images, videos, documents, emails, text messages and many more. Information is compressed so that it can lead to better and easier handling and faster and effective transmission. Nearly the entire internet uses multiple compression schemes to save time and costs involved in sharing information. Nowadays, the most frequently researched domain in computer science is comprised of images since imaging is an essential part of major industries of the world such as medicine, art, nature, wildlife, electronics, outer-space discoveries, etc. Scientists use images and imaging techniques to evaluate and study objects at the micro level and even those that are 55 million light years away. Most image compression techniques are usually of the form that take an image as an input, then perform a fixed set of operations on it, and give a compressed image as the output.

## II. Literature Review

### A. Artificial Neural Networks

Multiple traditional approaches besides machine learning are used to compress images such as transform coding, predictive coding, vector quantization, etc. All these techniques are used to achieve different outcomes: (1) transform coding is used to transform data into a form that takes up less space and is a good enough representative of the one it replaces, (2) predictive helps to remove elements that contribute to redundancy in an image, (3) vector quantization generates a codebook that works as a quantizer to compress the information in the image. All these techniques can even be combined with one another or multiple other methods to give what is called a hybrid approach to data compression.

The writers have described their work [1] by splitting it into various stages which are as follows: (1) image acquisition (2) segmentation of image into 2D images and conversion to 1D image that is given as input to the neural network (3) training of the ANN using Back propagation (4) quantizing the output of the bottleneck layer to achieve compression.

The reconstruction phase utilizes a receiver network that performs decompression of the compressed data obtained from the bottleneck. Both the compression and decompression stages of the proposed architecture are shown in Figure 1 and Figure 2 respectively.

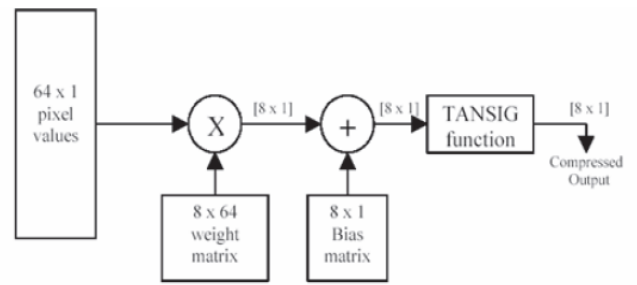

Fig. 1. Compression stage [1]

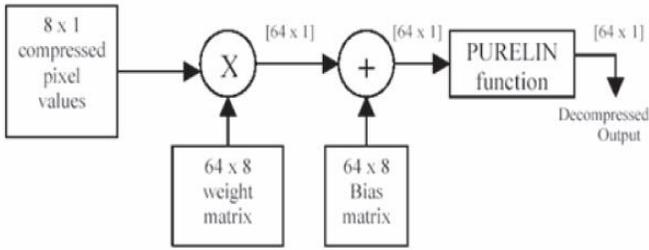

Fig. 2. Decompression stage [1]

For a particular image, the choice of transfer function and compression ratio used, yields different results. If the output of the decompression phase is better, the PSNR comes out to be on the higher end and the MSE low enough. Their work compares the result of using different compression ratios by plotting PSNR, MSE and Max Error where one of such plots is as shown in Figure 3.

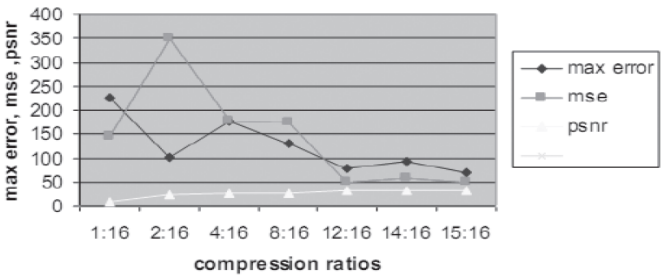

Fig. 3. Performance measure Graph [2]

### B. Convolutional Neural Nets

Image compression is broadly classified into 2 types: (1) lossy compression (2) Lossless compression. Lossy compression is used when the retainment of information in its entirety is not a requirement. It leads to larger image compression rates since data is lost. On the other hand, lossless image compression gives better visuals as it can reconstruct back the original image from the compressed image. Lossy compression systems are non-invertible whereas lossless are invertible. The work by [3], revolves around developing a new deep neural net architecture which implements a compression scheme for JPEG images and discusses ways to reduce the reconstruction loss made by Gaussian noise. The network trained here is adaptable to various types of images. The JPEG compression algorithm mainly has the following steps: (1) 8x8 block preparation, (2) Reshaping them into vectors of size 64x1, (3) Calculate discrete Fourier transform or discrete cosine transform of the vectors (4) Huffman encoding the result.

The algorithm described in [2] is comprised of two CNNs, one of which is used for encoding and the other for decoding. To reduce the artifacts caused by Gaussian noise, the author [] increased the magnitude of the noise so that the encoder output gives 0 or 1. Figure 4 shows the flow chart of the training CNN architecture.

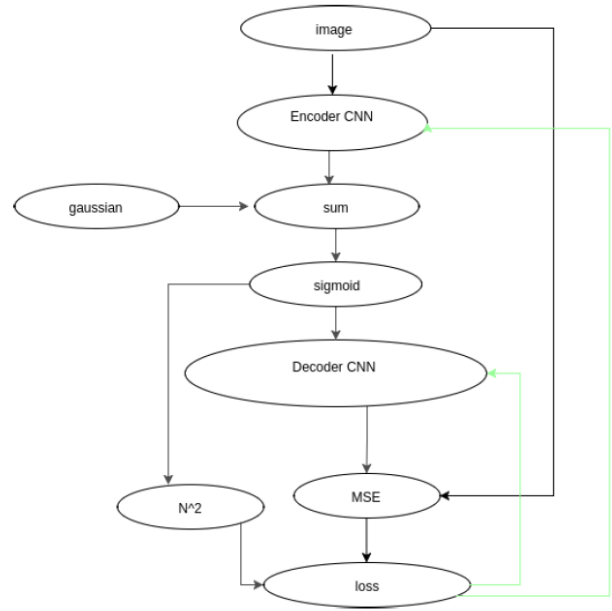

Fig. 4. Architecture of the training phase [2]

The input image, consisting of all 3 channels, is passed through the encoder CNN, Gaussian noise added to the output and the sigmoid function applied to get the encoded image data in binary form. The addition of the noise before application of sigmoid helps to encode the data in the form of 0s and 1s.

Therefore, the architecture presented here uses a deep neural network that works on JPEG images but can also be made to adapt to other image formats and yields a reduced construction loss and reduced artifacts. Other DNN image compression approaches [3] have shown improved PSNR values and faster convergence time.

### C. Hybrid Fuzzy Neural Network

A hybrid fuzzy neural approach is used [4] to compress and decompress an image. The process is similar to the function approximation problem where input-output relations are determined using given numerical data. The image compression architecture described here is designed as a function approximator. Compression is said to be achieved if the amount of space needed to store both the hidden unit values and connection weights of the decompressing layer is less than the space needed to store the original image information. The model is trained for multiple quantization bits and tested on different images.

Image compression is made possible due to multiple reasons: (1) redundancy in image data, (2) inability of human eye to detect small distortions in pixel values, (3) low resolution images sufficiently fulfil requirements, (4) limitations of the channel.

Performing image compression using neural networks has proved to be highly popular and efficient as shown by several authors [5] [6] [7] [8]. The work combines the benefits of both the fuzzy systems and neural nets to develop a hybrid system that has capabilities of both such as parallel processing, complex computations, ability to train on a variety of data, working with if-then fuzzy rules and membership functions. The model proposed [4] can work with both numeric and linguistic environments. Other hybrid approaches [9] also prove to be quite efficient at compressing images as they combine the strengths of two and more techniques.

Training images are preprocessed to give residual images which improve reconstruction since the average intensity of the testing image differs from the training images. This is done by first creating blocks of size $n \times n$ and calculating their mean. The residual blocks are then calculated by subtracting the quantized mean from the original blocks. Later, the mean of the quantized block is subtracted from the original pixel values to give the residual image. Finally, the fuzzy neural net is used to code the residual image, the architecture of which is show in Figure 5.

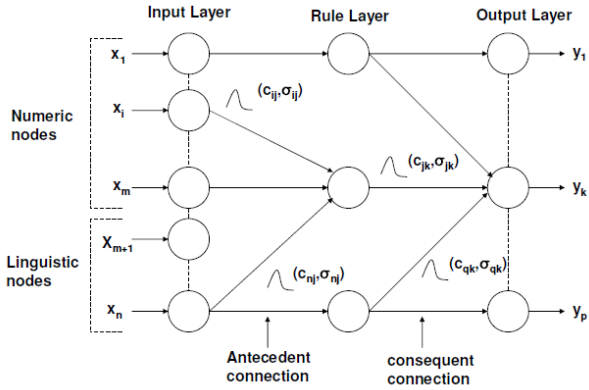

Fig. 5. FNN model archirecture [4]

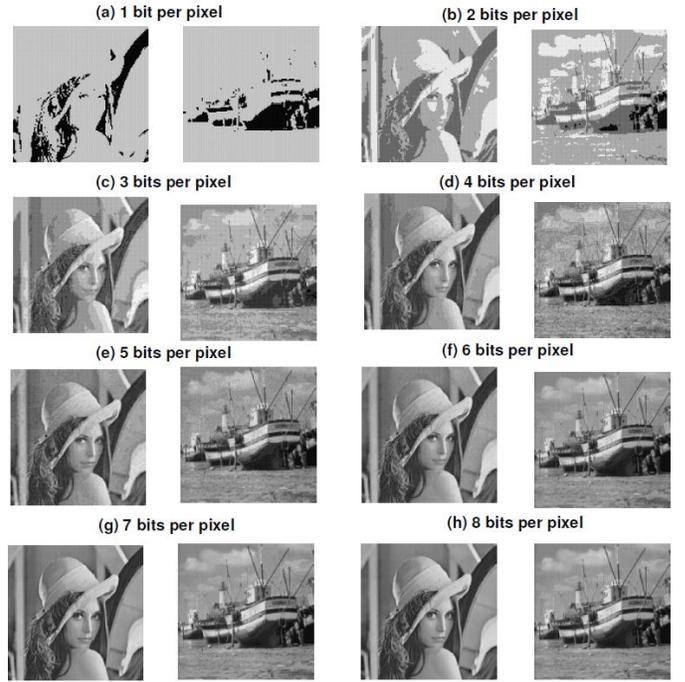

Fig. 6. Test images reconstructed using 8 different bits per pixel [5]

The model is designed in the following manner: (1) $x_1$ to $x_m$ and $x_{m+1}$ to $x_n$ are numeric and linguistic inputs to the model respectively, (2) each hidden node represents a rule, (3) each input-hidden connection is a fuzzy rule 'antecedent', (4) each hidden-output connection represents a fuzzy rule 'consequent', (5) center and spread of fuzzy weights from input nodes $i$ to rule nodes $j$, $w_{ij} = (c_{ij}, \sigma_{ij})$ (6) consequent fuzzy weights from rule nodes $j$ to output nodes $k$, $v_{jk} = (c_{jk}, \sigma_{jk})$, (7) use of mutual subsethood, a product aggregation operator at rule nodes, (8) use of volume defuzzification at output layer to generate numeric output $y_1$ to $y_p$, (9) training is performed using gradient descent technique.

The model was tested on the Lena image and the Boat image by varying the number of quantization bits as shown in Figure 6.

The values of parameters such as RMSE, PSNR and compression ratio were also calculated as shown in Table 1 [4]. For the Lena image, it can be seen that maximum PSNR is achieved at 8 bits per pixel whereas maximum compression is achieved at 1 bpp.

TABLE I. TEST IMAGE PARAMETERS

| Image | Quantization Bits | Rmse | PSNR (dB) | Compression Ratio |
|---|---|---|---|---|
| Lena | 1 | 69.764 | 11.258 | 30.56 |
| | 2 | 46.213 | 14.836 | 15.63 |
| | 3 | 27.765 | 19.261 | 10.50 |
| | 4 | 20.411 | 21.933 | 7.907 |
| | 5 | 17.301 | 23.368 | 6.340 |
| | 6 | 14.016 | 25.198 | 5.291 |
| | 7 | 10.708 | 27.537 | 4.541 |
| | 8 | 08.777 | 29.265 | 3.976 |
| Image | Quantization Bits | Rmse | PSNR (dB) | Compression Ratio |
| Boat | 1 | 61.897 | 12.297 | 30.56 |
| | 2 | 46.223 | 14.834 | 15.63 |
| | 3 | 31.283 | 18.225 | 10.50 |
| | 4 | 25.147 | 20.121 | 7.907 |
| | 5 | 21.121 | 21.636 | 6.340 |
| | 6 | 17.090 | 23.476 | 5.291 |
| | 7 | 14.791 | 24.731 | 4.541 |
| | 8 | 12.480 | 26.202 | 3.976 |

Similar behavior can be observed in the results for the Boat image. Hence it can be said that PSNR is directly related to Bits per pixel and Compression inversely related as can be observed in Figure 7 and Figure 8 which portray this relation along with comparing with a previously known compression model [10].

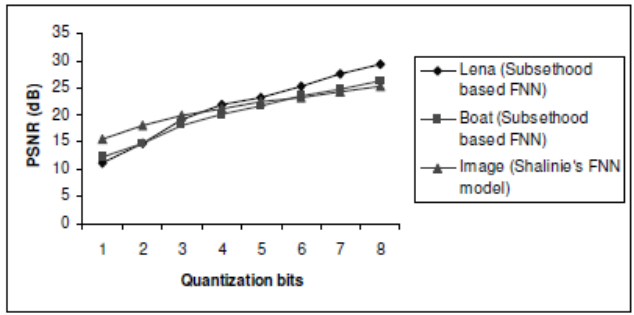

Fig. 7. PSNR vs. Quantization bits for Subsethood FNN(tested on Lena and Boat) and Shalinie's FNN [4]

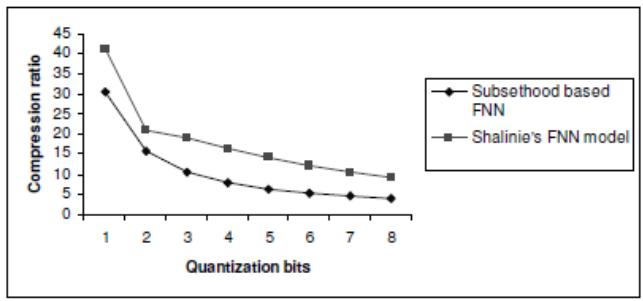

Fig. 8. Compression ratio vs. Quantization bits for Subsethood FNN and Shalinie's FNN [4]

*D. Genetic Algorithms*

The use of genetic algorithms to implement image compression schemes has become widespread as presented by the work on many authors [11] [12] [13].

Genetic algorithm is a technique that solves various optimization problems using the concept of natural selection that is derived from the biological evolution. It is used to solve for objective functions that are highly non-linear, stochastic, non-differentiable, and even discontinuous. This ability of GAs makes them highly beneficial to solve problems in electrical and electronics engineering, computer engineering, etc. A GA uses evolutionary operators such as natural selection, crossover and mutation iteratively to reach to the solution. Once the fitness, denoted by the objective function, of each population is calculated, the crossover and mutation operators are applied to mimic reproduction, and the process continues as shown in Figure 9, until the optimum is reached.

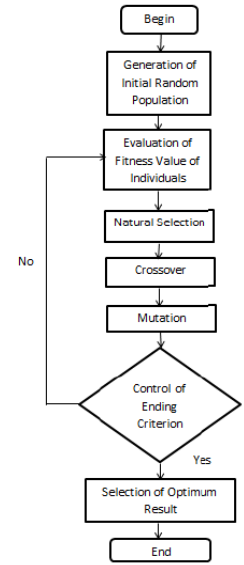

Fig. 9. Basic Genetic Algorithm Flowchart [14]

In the study research paper [14], the objective function is chosen to be MSE (Mean Square Error) where the initial image codebook is represented by random combinations.

The most representative and optimal codebook is generated using 3 different natural selection methods and 4 different codebook sizes: 8, 16, 32, and 64. Table 2 [14] depicts the results obtained for a codebook of size 64 and Figure 13 sows the decrease in MSE with increasing iterations.

TABLE II. GA RESUTS FOR 3 NATURAL SELECTION METHODS

| Lena | GA with Roulette Wheel Selection | GA with Elitist Selection | GA with Pool Based Natural Selection |
|---|---|---|---|
| Codeword Number (vector) | 1024 | 1024 | 1024 |
| Compression Ratio (bpp) | 0.0625 | 0.0625 | 0.0625 |
| MSE | 165.0547 | 165.0665 | 164.5081 |
| PSNR(db) | 25.9545 | 25.9542 | 25.9689 |

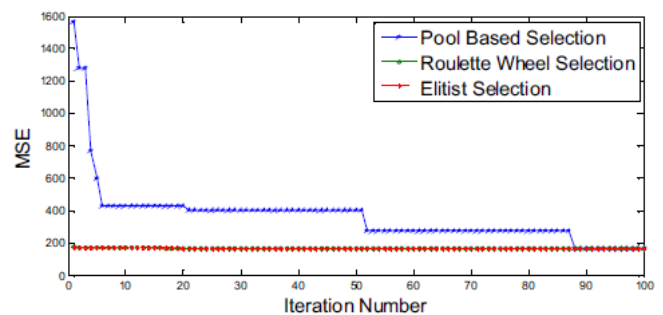

Fig. 10. MSE vs. Iteration number [14]

It can be seen from the results that Pool-based Natural selection exhibits the best performance compared to Roulette Wheel selection and Elitist Selection.

### III. PROPOSED METHOD

Genetic algorithms can be easily modified to solve various problem types and have lesser convergence time, thereby eliminating the need to craft special-purpose programs. In this work, a vector quantizer is designed using a genetic algorithm approach. The theory [15] [16] pertaining to the proposed algorithm is described in the sections that follow.

*A. Vector Quantization*

An image can be compressed by the process of quantization in 2 main ways: (1) Scalar Quantization: where each pixel is quantized individually depending on a fixed set of ranges, (2) Vector quantization: where a group of pixels are replaced with a group of best matching pixels from a codebook, the process is known as vector lookup.

Vector quantization is one of the emerging technologies in lossy data compression and is more widely used compared to scalar quantization. It helps to achieve significant compression ratio despite its simpler encoder and decoder design.

The codebook is designed through multiple techniques some of which are: K-means Algorithm, Lloyd Algorithm, Generalized Lloyd Algorithm and LBG (Linde-Buzo-Gray) Algorithm. The LBG algorithm uses either the: (1) Splitting Technique (the one used here) or (2) Pair-wise Nearest Neighbor (PNN) algorithm, to initialize the codebook. Each codebook entry represents a centroid.

The encoding phase also known as the training phase is comprised of:

1. Use training image to extract training vectors.
2. Specify size of codebook.
3. Initialize the codebook with the mean value of all training vectors.
4. Train codebook using the training vectors and functions such as *perturbcenter* and *recalculate*.
5. Save the generated codebook to an Excel document.

The decoding phase, which performs the vector lookup operation using the codebook generated in the training phase, also known as the testing phase, does the following tasks:
1. Load the testing image and process it into test vectors.
2. Load the previously saved codebook.
3. Use the *findmatch* function to find the best matching vector for each test vector from the codebook.
4. Reconstruct the Quantized image from the vectors and save the image.

*B. Functions*

1. 'perturbcenter':
Modifies the vector that is passed to it in a randomized manner. Done to implement the splitting technique.

2. 'recalculate':
Assigns training vectors to their corresponding centroids currently available in the codebook by calculating the error.
It then recalculates the centroids using the mean of all the training vectors that belong to each centroid.
3. 'findmatch':
Finds the best match for the test vector that is passed to it, from the codebook, using the error as a metric.

*C. Algorithm*

1. Initialize the number of maximum iterations, $T$ and $r=1$.
1. Preprocess training image to obtain $N$ training vectors,
$$X_i = \{x_1, x_2, x_3, ..., x_N\}$$

where $i=1$ to $N$ and each $x_i$ is a vector of size 4 x 1.
2. Initialize codebook with one vector which is the mean of $X$.
$$C_j = \{c_1 = mean_X\} \text{ where } j=1 \text{ to } M$$

3. Mutate all the existing elements of the codebook (done using *perturbcenter*). This leads to twice the number of centroids, the original ones along with the mutated version of each. Therefore, the number of centroids increase exponentially.
4. Calculate the error matrix for each $x_i$ w.r.t all the centroids.
$$e_{ij} = (x_i - c_j)'(x_i - c_j)$$

5. Update the membership vector $b$ of size N, with respect to the centroid to which each training vector belongs, that is, assign each training vector $x_i$ to the centroid $c_j$ for which it gives the least error.
$$b_{i=j} \ni \min e_{ij}$$

6. Perform migration (done using by calculating the mean of all training vectors belonging to each centroid and reassign the centroids to be equal to the mean.
$$c_j = mean(x_i) \ni b_{i=j}$$
7. $r=r+1$
8. Go to step 3 if $r < T$, else Stop.

*D. Results*

The codebook shown in Table 3 was obtained using the Lena image of size 100 x 100 pixel as shown in Figure 11.

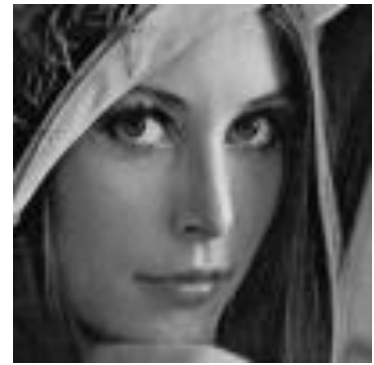

Fig. 11. Training Image

TABLE III.  64 WORD CODEBOOK

| 1 | 2 | 3 | 4 | 5 | 6 | 7 | 8 | 9 | 10 | 11 | 12 | 13 | 14 | 15 | 16 |
|---|---|---|---|---|---|---|---|---|---|---|---|---|---|---|---|
| 25.695 | 145.7111 | 76.55556 | 105.1231 | 47.96703 | 169.6667 | 168.8667 | 134.2544 | 19.60571 | 172.0526 | 85.42857 | 88.84615 | 28.13235 | 185.4348 | 64.41509 | 82.81818 |
| 26.38 | 145.5444 | 73.77778 | 103.0923 | 51.71429 | 172.2727 | 79.46667 | 135.2719 | 19.06286 | 175.7368 | 85.93333 | 86 | 27.38235 | 187.058 | 74.16981 | 118.2727 |
| 28.2 | 146.5778 | 86.88889 | 95.43077 | 40.73626 | 176.5758 | 72.06667 | 128.7193 | 20.13714 | 118.4211 | 81.72381 | 111.3077 | 46.20588 | 188.029 | 55.07547 | 127.0909 |
| 27.675 | 147.3667 | 113.2222 | 93.84615 | 38.94505 | 178.4242 | 41.2 | 131.3772 | 20.61143 | 127.5789 | 82.22857 | 115.8462 | 43.61765 | 185.6522 | 65.67925 | 162.2727 |

| 17 | 18 | 19 | 20 | 21 | 22 | 23 | 24 | 25 | 26 | 27 | 28 | 29 | 30 | 31 | 32 |
|---|---|---|---|---|---|---|---|---|---|---|---|---|---|---|---|
| 34.9455 | 152.353 | 57.4 | 112.227 | 69.9167 | 170.978 | 71.3571 | 124.011 | 15.0727 | 165.111 | 110.333 | 68.3636 | 21.8889 | 206.636 | 49.5 | 93.5455 |
| 32.9091 | 155.059 | 98 | 113.477 | 40.25 | 172.217 | 97.6429 | 121.57 | 14.7273 | 163.222 | 82.619 | 70.4545 | 51.4444 | 205.545 | 48.6842 | 109.455 |
| 21.1091 | 159.176 | 109.6 | 114.188 | 48.3333 | 160.391 | 42 | 128.151 | 12.0182 | 151.667 | 96.0476 | 141.545 | 27.6667 | 212.273 | 71 | 157.273 |
| 23.8545 | 166.912 | 100.2 | 113.922 | 25.6667 | 162.457 | 53.4286 | 126.108 | 11.8727 | 126.667 | 69.1429 | 145.545 | 67.1111 | 214.455 | 60.5789 | 165.545 |

| 33 | 34 | 35 | 36 | 37 | 38 | 39 | 40 | 41 | 42 | 43 | 44 | 45 | 46 | 47 | 48 |
|---|---|---|---|---|---|---|---|---|---|---|---|---|---|---|---|
| 35.4038 | 148.588 | 52.1667 | 98.9434 | 43.58 | 168.182 | 120.412 | 149.19 | 13.6 | 184.889 | 87.2041 | 113.556 | 33.7436 | 196.545 | 58.6111 | 73.25 |
| 27.3462 | 161.059 | 62.2222 | 100.245 | 42.04 | 170.364 | 100.235 | 136.381 | 17.4 | 165.444 | 90.2449 | 92.4444 | 36.1538 | 199.818 | 62.3056 | 133 |
| 33.9808 | 157.235 | 82.7778 | 103.009 | 29.94 | 185.364 | 53.5294 | 128 | 32.4 | 148.556 | 88.9898 | 128.444 | 63.2308 | 183.273 | 67.9167 | 102.625 |
| 29.7885 | 152 | 90.5556 | 102.934 | 35.76 | 161 | 54.7647 | 111.286 | 25 | 89.8889 | 93.9796 | 107.778 | 62.2821 | 189.545 | 67.6111 | 153.25 |

| 49 | 50 | 51 | 52 | 53 | 54 | 55 | 56 | 57 | 58 | 59 | 60 | 61 | 62 | 63 | 64 |
|---|---|---|---|---|---|---|---|---|---|---|---|---|---|---|---|
| 28.1429 | 142.412 | 32.1667 | 124.061 | 76 | 190 | 97.2222 | 115.167 | 11 | 169.5 | 94.8 | 54.5882 | 18.6667 | 195.105 | 62.9474 | 127.5 |
| 40.4643 | 133.706 | 58.6667 | 122.727 | 46 | 171.167 | 99.3333 | 115.278 | 9.5 | 148.571 | 57.5 | 42.4118 | 27.7222 | 194.842 | 37.4737 | 102.5 |
| 16.8929 | 166.471 | 111.5 | 97.6061 | 20.4762 | 164.333 | 32.4444 | 133.444 | 7 | 155 | 86.2 | 131.294 | 46 | 200.632 | 70.5789 | 153.5 |
| 21.2857 | 158.647 | 112 | 96.5455 | 18.6667 | 155.5 | 36.8889 | 134.222 | 8 | 141.929 | 69.1 | 132.118 | 82.5 | 199.316 | 50.4211 | 156.5 |

The results obtained while compressing a 200 x 200 pixel image using a codebook of size 64 are shown in Figure 12 where the image on the left is the original image and the one on the right is the quantized image.

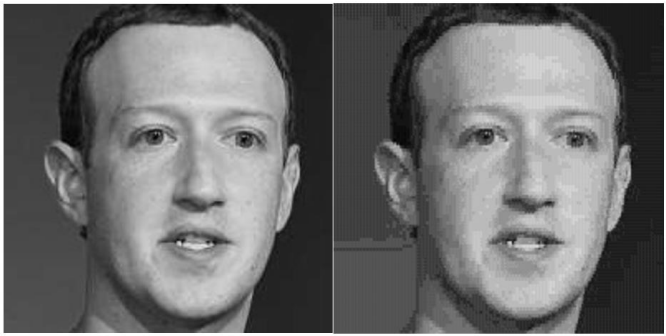

Fig. 12. Experimental results

*E. Performance Evaluation Data*

- Time to generate 64-word codebook from 2500 training vectors = 0.564292 seconds
- Time taken to quantize the 200 x 200 testing image using 64 level codebook= 0.839802 seconds
- PSNR = 29.3415
- Shannon Entropy= 4.6031
- New Bitrate = 4.64
- Original bitrate = 8

## IV. CONCLUSION

After reviewing a number of image compression schemes, it can be concluded that a compression system with fixed complexity produces lossy compression for larger images whereas it tends to be less lossy for smaller images. Better reconstruction is achieved by more complex systems but by using simpler systems like smaller neural nets, the compression becomes faster and efficient by compromising on visual quality.

## V. FUTURE WORK

The algorithm can be modified to incorporate multiple other features of genetic evolution such as the crossover operator and the test of convergence along with a fixed maximum number of iterations. The work presented here was bound by time constraints and hence is limited in its scope.